\documentclass[a4paper,10pt,onecolumn,review,3p,authoryear]{elsarticle}

\usepackage{amssymb}
\usepackage{amsmath}
\usepackage{graphicx}
\usepackage{url}
\usepackage{subfig}

\newcommand{\repositoryURL}{\url{https://github.com/wur-abe/uav_adaptive_planner}}

\journal{Smart Agricultural Technology}

\begin{document}

\makeatletter
\def\ps@pprintTitle{%
  \let\@oddhead\@empty
  \let\@evenhead\@empty
  \let\@oddfoot\@empty
  \let\@evenfoot\@oddfoot
}
\makeatother

\begin{frontmatter}

\title{Adaptive path planning for efficient object search by UAVs in agricultural fields}

\author[1]{Rick van Essen}\corref{cor1}
\ead{rick.vanessen@wur.nl}
\author[1]{Eldert van Henten}
\author[2]{Lammert Kooistra}
\author[1]{Gert Kootstra}

\affiliation[1]{
    organization={Agricultural Biosystems Engineering, Department of Plant Sciences, Wageningen University and Research, 6700 AA},
    city={Wageningen},
    country={The Netherlands}
}
\affiliation[2]{
    organization={Laboratory of Geo-information Science and Remote Sensing, Department of Environmental Sciences, Wageningen University and Research, 6700 AA},
    city={Wageningen},
    country={The Netherlands}
}
\cortext[cor1]{Corresponding author.}

\begin{abstract}
This paper presents an adaptive path planner for object search in agricultural fields using UAVs. The path planner uses a high-altitude coverage flight path and plans additional low-altitude inspections when the detection network is uncertain. The path planner was evaluated in an offline simulation environment containing real-world images. We trained a YOLOv8 detection network to detect artificial plants placed in grass fields to showcase the potential of our path planner. We evaluated the effect of different detection certainty measures, optimized the path planning parameters, investigated the effects of localization errors, and different numbers of objects in the field. The YOLOv8 detection confidence worked best to differentiate between true and false positive detections and was therefore used in the adaptive planner. The optimal parameters of the path planner depended on the distribution of objects in the field. When the objects were uniformly distributed, more low-altitude inspections were needed compared to a non-uniform distribution of objects, resulting in a longer path length. The adaptive planner proved to be robust against localization uncertainty. When increasing the number of objects, the flight path length increased, especially when the objects were uniformly distributed. When the objects were non-uniformly distributed, the adaptive path planner yielded a shorter path than a low-altitude coverage path, even with a high number of objects. Overall, the presented adaptive path planner allowed finding non-uniformly distributed objects in a field faster than a coverage path planner and resulted in a compatible detection accuracy. The path planner is made available at \repositoryURL.
\end{abstract}


\begin{keyword}
Adaptive path planning \sep Object detection \sep Detection certainty \sep Unmanned aerial vehicles \sep Drones
\end{keyword}

\end{frontmatter}


\section{Introduction}
Over the last years, the use of Unmanned Aerial Vehicles (UAVs) in agriculture has grown considerably \citep{Rejeb2022}. Applications are ranging from yield estimation \citep{Yuan2024} to weed \citep{Murad2023} and disease detection \citep{Shahi2023_review, Kouadio2023}. These applications require objects, such as weeds and diseases, to be detected in a large field. Typically, the UAV follows a pre-defined low-altitude coverage flight path at fixed altitude \citep{Cabreira2019}. These paths are easy to use and cover the whole field with an equal spatial resolution. However, for an object detection task, it is not needed to cover the whole field with an equal spatial resolution. Flying at a higher altitude will increase the field-of-view of the UAV and thereby decrease the total flight path length and associated flight time, but decrease the spatial resolution and thereby the detection accuracy. Only decreasing the altitude of the UAV when the detection is uncertain can decrease the flight time while achieving a similar detection as a low-altitude coverage path. To optimize the battery capacity of UAVs, it is important to use a method that minimizes the flight path length while maintaining a good detection accuracy. Especially when using UAVs for autonomous and frequent monitoring of fields, it is important to limit the flight time to increase the area that can be monitored using a single UAV.

Adaptive path planning, sometimes called informative path planning, is a path planning approach that aims to plan a trajectory that maximizes gathering of task-relevant information while keeping within specific resource limits, such as battery capacity \citep{Popovic2024}. This can be done by making the path planning reactive to real-time data, making sequential decisions about the UAV's flight path based on newly gathered information \citep{Popovic2024}. Adaptive path planning with UAVs is used, for instance, to detect anomalies in ocean topography \citep{Blanchard2022}, detection of weeds in agricultural fields \citep{Popovic2017}, and multi-resolution semantic segmentation \citep{Stache2023}. \citet{Popovic2017} showed that an adaptive path planner yielded a 50\% lower map uncertainty (measured using entropy) in the same amount of time for active weed classification (simulated using printed markers) compared to a low-altitude coverage flight path. \citet{Stache2023} proposed a path planner for semantic segmentation that uses an estimated relation between altitude and segmentation accuracy. It performs close inspections at lower altitudes only when required and thereby minimizing the flight time. The path planner was demonstrated by segmenting soil, plants, and weeds by artificially flying over the WeedMap dataset and elements like asphalt, buildings, water, etc., by artificially flying over satellite images. They show a high segmentation accuracy while minimizing the flight time. \Citet{vanEssen2024} presents a learning based adaptive path planner that uses Reinforcement Learning to localize objects of interest in a field. The learned policy uses uncertain prior knowledge of the field together with object detections in the current field-of-view to control a UAV. In a simulated environment, they show that learned path outperforms a coverage planner by localizing 80\% of the objects in less than half of the flight time. However, all these methods do not directly use the certainty estimation by the detection or segmentation network, even though such an estimate could be valuable for decision making in an adaptive path planner.


When increasing the altitude of the UAV, the field-of-view increases, and thereby the spatial resolution of the image decreases. Due to a smaller number of pixels describing an object, the number of false positive and false negative detections will increase, and thereby the accuracy decreases. An adaptive path planner can exploit the trade-off between altitude and accuracy by only capturing low-altitude images when needed. For example, at a high altitude, the detection network may detect an object that appears to be a weed. Based on the uncertainty associated with the detection, a decision can be made to accept the detection as reliable or gather a new image at a lower altitude. 

A detection network typically outputs a confidence belonging to a detection as a measure of uncertainty. When the detection confidence of an object is low, an adaptive path planner can, for example, plan a low-altitude inspection of this object. However, neural networks tend to over- or under-estimate the confidence \citep{Gawlikowski2023}, leading to a rejection of correct detections and an inclusion of wrong detections. Several methods have been applied in literature to provide a better uncertainty estimation. Applying Monte-Carlo Dropout (MCD) during inference is often applied due to its low implementation effort \citep{Gawlikowski2023}. \citet{Myojin2019} showed a 1-2\% increase in accuracy when using uncertainty measures estimated using MCD for detecting lunar craters on images using an adapted version of YOLOv3. In agriculture, MCD has been applied for uncertainty estimation in active learning for broccoli detection \citep{Blok2022}, fish species detection \citep{Sokolova2024}, obstacle detection in solar farms \citep{Rodriguez-vazquez2024}, and assessing uncertainty in chicken plumage condition \citep{Lamping2023}. They illustrate that MCD can be used to estimate uncertainty in neural networks.

The objective of this paper is to develop and evaluate an adaptive path planner that finds objects in agricultural fields faster than a low-altitude coverage path planner while maintaining a similar detection performance. This adaptive path planning method follows a high-altitude coverage flight path and uses the detection certainty to perform low-altitude inspections when the detection certainty is low. Compared to existing approaches in the literature, the developed adaptive path planner directly uses uncertainty estimates from an object detection network, making it more flexible for other applications. Besides the field boundaries, it does not need prior knowledge of the field, and the performance was evaluated on real-world data in combination with a real detection network. Furthermore, we open-sourced the code at \repositoryURL. Specifically, we (1) examined the effect of different object detection certainty estimation methods calculated using MCD, (2) optimized the flight altitude and certainty thresholds of the planner, (3) analyzed the influence of errors in the UAV's localization, and finally, (4) investigated the efficiency of the adaptive path planner to the number of objects in the field. 


\section{Materials and methods}
In this section, we describe the adaptive path planner (section \ref{sec:adaptive_path_planner}), the detection uncertainty estimation methods (section \ref{sec:detection_certainty_measures}), the data collection (section \ref{sec:data_collection}), the training procedure for the detection network (section \ref{sec:detection_network_training}), the offline simulation environment (section \ref{sec:offline_sim_env}), the baseline (section \ref{sec:baseline}), the performance metrics (section \ref{sec:performance_metrics}) and finally the experiments (section \ref{sec:experiments}). 

\subsection{Adaptive path planner}
\label{sec:adaptive_path_planner}
\begin{figure}[!ht]
   \centering
   \includegraphics[width=\textwidth]{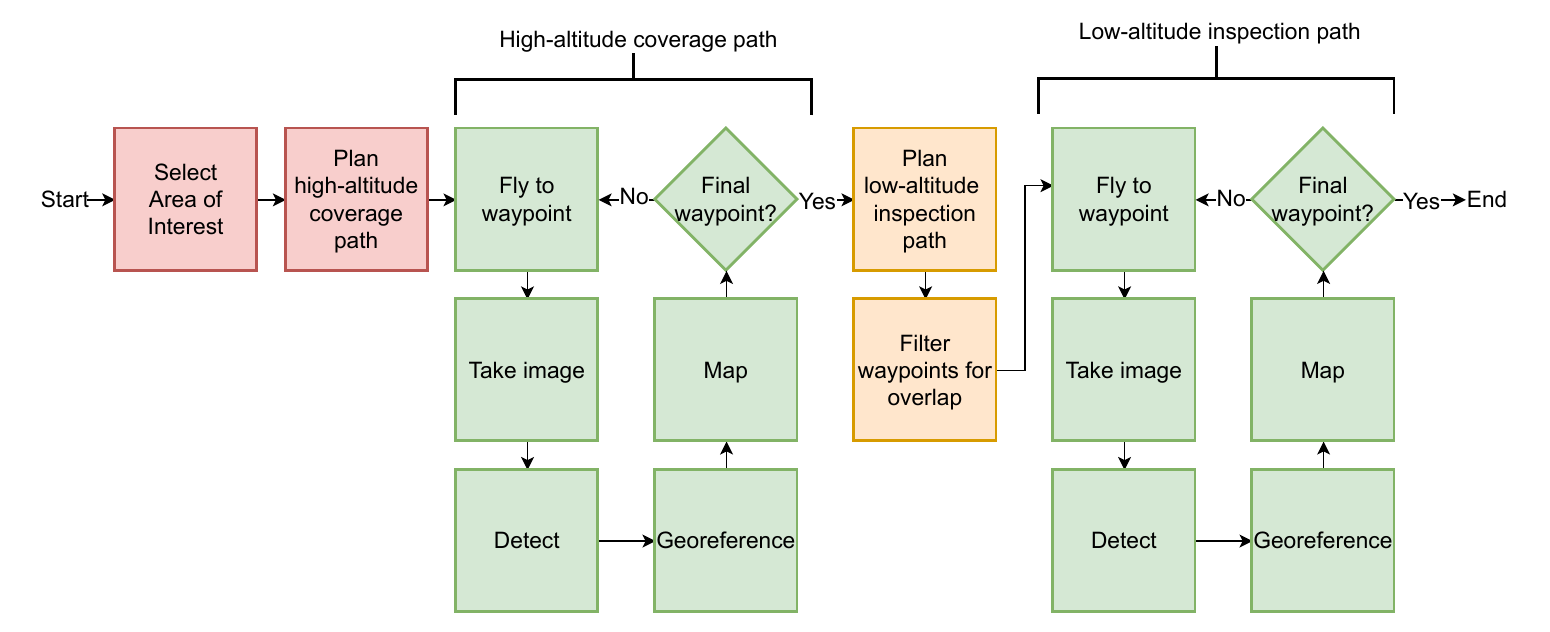}
   \caption{Overview of the different parts of the adaptive path planner: planning the high-altitude coverage path (red, see section \ref{sec:coverage_flight_path}), planning the low-altitude inspection flight path (orange, see section \ref{sec:inspection_flight_path}) and executing the flight paths (green).}
   \label{fig:planner_flowchart}
\end{figure}

The concept of the proposed adaptive path planner is visualized in Figure \ref{fig:planner_flowchart}. First, the flight area to be covered by the UAV needs to be manually selected. Based on the flight area, a high-altitude coverage path is planned (section \ref{sec:coverage_flight_path}) and executed. At each waypoint, we take an image from a top-down perspective (nadir) and a detection network is used to detect the objects in real-time (section \ref{sec:detection}). The detected objects are then georeferenced (section \ref{sec:georeferencing}) and mapped (section \ref{sec:mapping}). After completing the high-altitude coverage path, an inspection path is planned with waypoints at a lower altitude at the locations of the object detections that were uncertain (section \ref{sec:inspection_flight_path}). To limit the number of waypoints for low-altitude inspection, the waypoints are filtered to remove overlapping waypoints. The inspection flight path is executed in a similar way as the high-altitude coverage path. During the inspection flight, the mapped objects are updated and previously unseen objects are added when needed. Finally, all object locations in the map (both from the high- and low-altitude flight) with a certainty larger than $c_\textrm{eval}$ are used for evaluation. 

\subsubsection{Planning of high-altitude coverage path}
\label{sec:coverage_flight_path}
\begin{figure} 
   \centering
   \includegraphics[width=\textwidth]{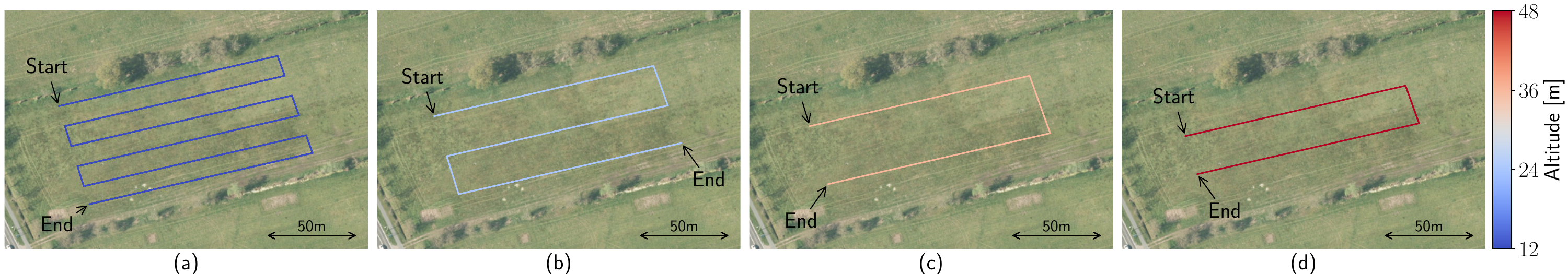}
   \caption{Example of the coverage flight path for 12m (a), 24m (b), 36m (c), and 48m(d) altitude.}
   \label{fig:baseline_flight_paths}
\end{figure}

The first step of the proposed method is to manually select a flight area by drawing a polygon on a satellite image indicating the boundaries of the field. Within this area, a coverage flight path is planned using Fields2Cover \citep{Mier2023} at an altitude of $h_\textrm{cov}$. The field-of-view, $\textrm{FOV}(h)$ (m) at altitude $h$ is calculated by:

\begin{equation}
    \label{eq:fov}
    \textrm{FOV}(h) = \frac{\textbf{c} \odot h}{f},
\end{equation}
\noindent
where $\odot$ denotes an element-wise multiplication, $\textbf{c} = [c_w, c_h]^T$ the sensor width and height in mm, and $f$ the camera focal length in mm. However, when the localization of the UAV is imperfect, taking the calculated field-of-view as row width for the coverage path planner would then result in missing a part of the field. Therefore, to reduce the impact of localization errors, we set the row width to 90\% of the field-of-view by adding an overlap of 5\% of the field-of-view to both sides. Likewise, the distance between two waypoints on the flight path is set to 90\% of the field-of-view. Figure \ref{fig:baseline_flight_paths} shows an example of the coverage flight paths for 12m, 24m, 36m, and 48m altitude. 

\subsubsection{Detection}
\label{sec:detection}
At each waypoint, an image is taken from a top-down perspective. In this image, objects are detected using YOLOv8. \citep{Jocher2023}. YOLOv8 is the eighth version of the YOLO (You Only Look Once) family of fully convolutional neural networks for object detection, which predicts both the bounding box coordinates as well as the class probabilities in a single pass through the network. The nano version of YOLOv8 is used, which was selected to have low hardware requirements, and allows for detection on board a UAV. During inference, an image input size of 2048x2048 pixels is used, and the minimum confidence is set to $c_\textrm{reject}$.

\subsubsection{Georeferencing}
\label{sec:georeferencing}
Georeferencing object detections is done in real-time using the UAV's position and heading. The first step is to calculate the field-of-view, $\textrm{FOV}(h)$, at the current altitude of the UAV using Equation \ref{eq:fov}. The pixel coordinates of the detected object $o$ in the image $[u_o, v_o]^T$ (px) can be transformed in the local camera frame, $[x^\textrm{local}_o,y^\textrm{local}_o]^T$ (m), by:

\begin{equation}
   \begin{bmatrix}
        x^\textrm{local}_o \\
        y^\textrm{local}_o
    \end{bmatrix}
    =
    \Biggl(
    \begin{bmatrix}
        u_o \\
        v_o
    \end{bmatrix}
    \odot
    \begin{bmatrix}
        \frac{1}{I_w} \\
        \frac{1}{I_h}
    \end{bmatrix}
    - \frac{1}{2}
    \Biggr)
    \odot 
    \textrm{FOV}(h),
\end{equation}
\noindent

where $I_w$ and $I_h$ are the width and height of the image in pixels. Finally, the local coordinates can be transformed to the world Universal Transverse Mercator (UTM) coordinates, $[x^\textrm{utm}_o,y^\textrm{utm}_o]^T$ (representing Easting and Northing), using an affine rotation:

\begin{equation}
    \label{eq:world_coordinates}
    \begin{bmatrix}
        x^\textrm{utm}_o \\
        y^\textrm{utm}_o
    \end{bmatrix}
    = 
    \begin{bmatrix}
        \cos{\psi} & -\sin{\psi} \\
        \sin{\psi} & \cos{\psi}
    \end{bmatrix}
    \cdot
    \begin{bmatrix}
        x^\textrm{local}_o \\
        -y^\textrm{local}_o
    \end{bmatrix}
    + 
    \begin{bmatrix}
        x^\textrm{utm}_\textrm{uav} \\
        y^\textrm{utm}_\textrm{uav}
    \end{bmatrix}
    ,
\end{equation}
\noindent
where $\psi$ is the counter-clockwise heading of the drone relative with respect to the north and $[x^\textrm{utm}_\textrm{uav},y^\textrm{utm}_\textrm{uav}]^T$ the UAV's UTM coordinates. Since UTM coordinates use the North-up orientation, we take the negative of $y^\textrm{local}_o$.

\subsubsection{Mapping}
\label{sec:mapping}
The GNSS location of each detection is stored in a map during operation. Because of overlap between waypoints and flying at different altitudes, objects may be observed multiple times. To merge these observations, all detections with an Euclidean distance of less than $d_{\textrm{dist}}$ are considered the same object. The default value of $d_\textrm{dist}=0.35$m was selected based on the minimum distance between the objects in the field. 

If an object from the map was not detected in the current field-of-view while it should have been visible, the mapped object is assumed to be a false positive and removed from the map when both conditions apply:

\begin{itemize}
    \item \textbf{If} the current altitude is lower than the minimum altitude of all observations of that object,
    \item \textbf{And} the highest observed certainty is lower than $c_\textrm{accept}$.
\end{itemize}

\noindent
Merging a detection in the map is done according to the following rules:

\begin{itemize}
    \item \textbf{If} the altitude of the UAV associated with the new detection is lower than the minimum altitude of all previous observations of that mapped object, overwrite the certainty, the class name and the location by the new detection,
    \item \textbf{Else If} the certainty of the new detection is higher than the previously stored object location, overwrite the certainty, the class name and the location by the new detection
\end{itemize}

\subsubsection{Planning of low-altitude inspection flight path}
\label{sec:inspection_flight_path}
\begin{figure}[!ht]
   \centering
   \includegraphics[width=0.4\textwidth]{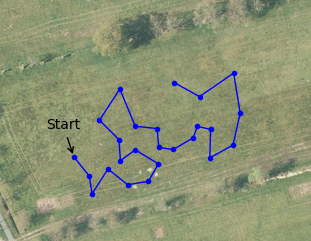}
   \caption{Example of an inspection flight path that visits uncertain detections to capture an image at the indicated waypoints (blue dots) with a higher spatial resolution. The resulting low-altitude inspection flight path is indicated with the blue line.}
   \label{fig:example_inspection_flight_path}
\end{figure}

After executing the high-altitude coverage flight path, an inspection flight path is planned to observe objects in the map with a low certainty (see section \ref{sec:detection_certainty_measures}) in more spatial detail at a lower altitude. For all the objects stored in the map, we make a decision $\mathcal{D}_i$ for each object $i$, to accept, reject, or have a closer inspection of the object using:

\begin{equation}
  \mathcal{D}_i =
    \begin{cases}
      \text{accept} & c_i > c_{\textrm{accept}} \\
      \text{reject} & c_i < c_{\textrm{reject}} \\
      \text{inspect} & \text{otherwise}
    \end{cases},       
\end{equation}
\noindent
where $c_i$ is the certainty corresponding to the object (see section \ref{sec:detection_certainty_measures}), $c_{\textrm{accept}}$ the certainty threshold to accept an object without further inspection and $c_{\textrm{reject}}$ the certainty threshold to reject and ignore an object. An inspection flight path is planned at an altitude of $h_\textrm{inspect}$ to visit all objects where $\mathcal{D}_i$ equals 'inspect'. The resulting list of waypoints is ordered to get the shortest route using iterative stochastic local search by minimizing the total distance \citep{Mulvad2022}. For every waypoint, the heading is set towards the next waypoint. To shorten the created path, the waypoints are iteratively filtered by removing next waypoints when the corresponding detection is expected to be visible in the field-of-view of the current waypoint. Figure \ref{fig:example_inspection_flight_path} shows an example of an inspection flight path.

\subsection{Detection certainty measures}
\label{sec:detection_certainty_measures}
As described above, the adaptive path planner bases decisions on the certainty of the detection network to inspect uncertain detections that are possibly incorrect. This requires a correlation between the certainty and errors made by the detection network, since the certainty is used as a probability of whether an object is present or not. In this paper, we evaluate six different methods to calculate uncertainty. The first one is the standard detection certainty as returned by the detection network for the best scoring class,  $c^\textrm{yolo}$. This value represents a combination of the certainty for the bounding box size and location, and the class of an object. However, most networks tend to over- or under-estimate the confidence \citep{Gawlikowski2023}. Therefore, we calculate five additional certainty measures using Monte-Carlo dropout (MCD). Calculating certainties using MCD is done in two steps: applying MCD on the YOLOv8 network (section \ref{sec:mcd}), and calculating the certainty measures (section \ref{sec:unc_calc}).

\subsubsection{Monte-Carlo dropout}
\label{sec:mcd}
Dropout is a regularization technique in neural networks that randomly drops connections between neurons, and is regularly applied to avoid over-fitting during training by forcing the network to learn different 'routes' through the network to get to the same outcome \citep{Srivastava2014}. \citet{Gal2016} introduced Monte-Carlo dropout during inference and showed that it can be used to assess model uncertainty. Each connection where dropout is applied has the probability of $q$ to be removed, and this process is repeated for $N$ runs in the network. When the network produces the same output during all runs while randomly removing some connections, it indicates that the network is certain. However, when every run results in a different output, the model is uncertain how to handle the input.

\begin{figure}[!ht]
   \centering
   \includegraphics[width=\textwidth]{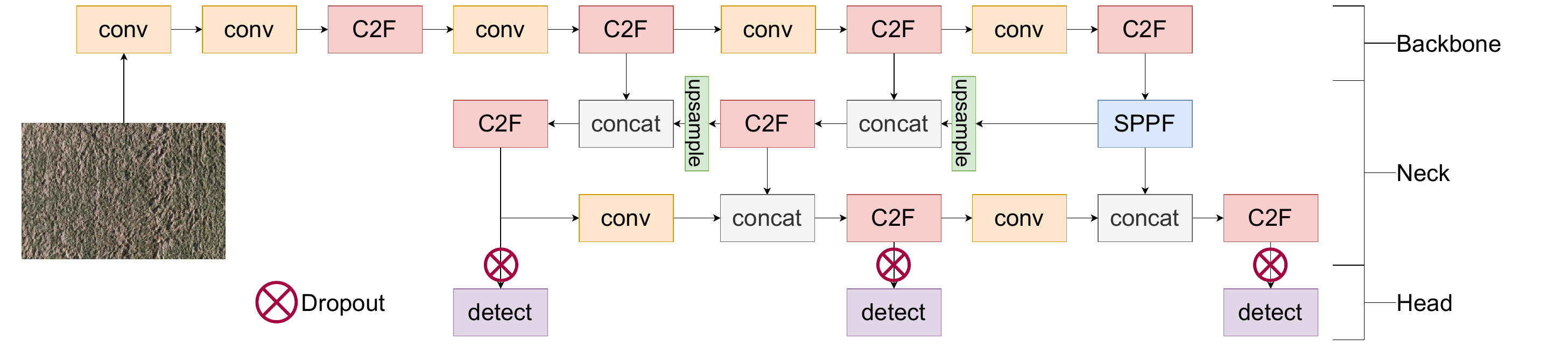}
   \caption{Schematic (simplified) overview of the YOLOv8 architecture with the locations indicated where Monte-Carlo dropout is applied between the neck and head of the network. Each ‘conv’ block contains a 2D convolution layer, batch normalization, and a sigmoid activation function. A ‘C2F’ block is a Cross-Stage Partial block with two convolutional layers. ‘SPPF’ indicates a Spatial Pyramid Pooling block, and ‘concat’ is a concatenation layer. The ‘detect’ block includes two ‘conv’ blocks followed by a 2D convolution layer to create the list of detections. See \citet{Jocher2023} for more details.}
   \label{fig:yolov8_with_dropout}
\end{figure}

Figure \ref{fig:yolov8_with_dropout} shows the architecture of YOLOv8, indicating the backbone, the neck and head of the network, and the several connections that pass through information from earlier layers in the network. The locations where dropout is applied are indicated with crosses. The dropout is applied between the neck and the head of the network, which enables us to reuse the output of the backbone and neck during inference and save computational time. 

For an image $i$, we use one forward pass without dropout, $D_i^\textrm{ref}$, to provide the network's prediction and $N=20$ passes with a dropout probability of $q=0.25$. These detections are combined in a detection set $S_i$:

\begin{equation}
     S_i=\{D_{1,i},D_{2,i},...,D_{N,i}\},
\end{equation}
\noindent
where $D_{j,i}$, is the detection of object $i$ in the MCD run $j$. $D_{j,i}$ contains the bounding box coordinates, class probabilities, and confidence score. To combine the MCD detections with a the reference detection $D_i^\textrm{ref}$ for each forward pass $j$, we calculate the intersection-over-union (IoU) with all MCD detections $D_{j,i}^\textrm{mc}$:

\begin{equation}
    \textrm{IoU}(D_i^\textrm{ref}, D_{j,i}^{\textrm{mc}}) = \frac{|D_i^{\textrm{ref}} \cap D_{j,i}^{\textrm{mc}}|}{|D_i^{\textrm{ref}} \cup D_{j,i}^{\textrm{mc}}|}.
\end{equation}
\noindent
When the IoU is larger than 0.5, $D_{j,i}^\textrm{mc}$ is added to the detection set $S_i$. Note that combining the reference detection with the MCD detections is only based on their bounding box coordinates and is class-independent.

\subsubsection{Certainty calculation}
\label{sec:unc_calc}
Next to the standard detection certainty, we calculate five additional certainty measures per object detection based on the detection sets: the average YOLO confidence, occurrence certainty, the localization certainty, the class certainty and a combined certainty.

\noindent
\textbf{Average YOLO certainty:}\\
The average YOLO certainty, $c_i^\textrm{yolo}$ of reference detection $i$ is calculated by taking the mean of the YOLO confidence values corresponding to the detection set $S_i$:

\begin{equation}
    c_i^\textrm{yolo} = \frac{1}{\|S_i\|} \cdot \sum_{j=1}^{\|S_i\|} c_{j,i}^\textrm{yolo}, 
\end{equation}
\noindent
where $c_{j,i}^\textrm{yolo}$ is the YOLOv8 confidence value belonging to detection $j$ in the detection set. When $S_i$ is empty, $c_i^\textrm{yolo}$ is set to 0.0.

\noindent
\textbf{Occurrence certainty:}\\
The occurrence certainty, $c_i^\textrm{occ}$ of reference detection $i$ is calculated by:

\begin{equation}
    c_i^\textrm{occ} = \frac{\|S_i\|}{N}.
\end{equation}

\noindent
\textbf{Location certainty:}\\
The location certainty, $c_i^\textrm{loc}$ of reference detection $i$, is the average IoU of detections in detection set $S_i$ compared to the reference detection:

\begin{equation}
    c_i^\textrm{loc} = \frac{1}{\|S_i\|} \cdot \sum_{D_{j,i}^\textrm{mc} \in S_i} \textrm{IoU} (D_i^\textrm{ref}, D_{j,i}^\textrm{mc}). 
\end{equation}
\noindent
The localization certainty is based on the work of \citet{Blok2022}. When $S_i$ is empty, $c_i^\textrm{loc}$ is set to 0.0.

\noindent
\textbf{Class certainty:}\\
The class certainty of reference detection $i$ is calculated using the entropy, $H_i$:

\begin{equation}
    H_i(k) = -\sum_{j=1}^{\|S_i\|}P_{j,i}(k) \cdot \ln P_{j,i}(k),
\end{equation}
\noindent
where $P_{j,i}(k)$ is the class probability for class $k \in K$ belonging to detection $j$ in the detection set $S_i$. The entropy values are normalized to be in $[0, 1]$ using the softmax function, $\sigma_i(k)$:

\begin{equation}
    \label{eq:softmax}
    \sigma_i(k) = \frac{e^{H_i(k)}}{\sum\limits_{k_z \in K} e^{H_i(k_z)}}.
\end{equation}
\noindent
Finally, we calculate the class certainty, $c_i^\textrm{cls}$, by taking the maximum normalized certainty across all classes:

\begin{equation}
    c_i^\textrm{cls} = \max_{k \in K}\bigl(\sigma_i(k)\bigr).
\end{equation}
\noindent
The class probabilities $P(k)$ are calculated by using the softmax function (Equation \ref{eq:softmax}) on the class activation values from the last layer of the detection network.

\noindent
\textbf{Combined certainty:}\\
The combined certainty, $c_i^\textrm{comb}$ of reference detection $i$, is calculated by taking the product of the occurrence, location, and class certainty:

\begin{equation}
    c_i^\textrm{comb} = c_i^\textrm{occ} \cdot c_i^\textrm{loc} \cdot c_i^\textrm{cls}.
\end{equation}
\noindent
Detections with high occurrence, location, and class certainty will have a high combined certainty, whereas a lower value in any of these certainties will result in a lower combined certainty.

\subsection{Data collection}
\label{sec:data_collection}
\begin{figure}[!ht] 
    \centering
    \subfloat[]{
        \includegraphics[width=0.15\linewidth]{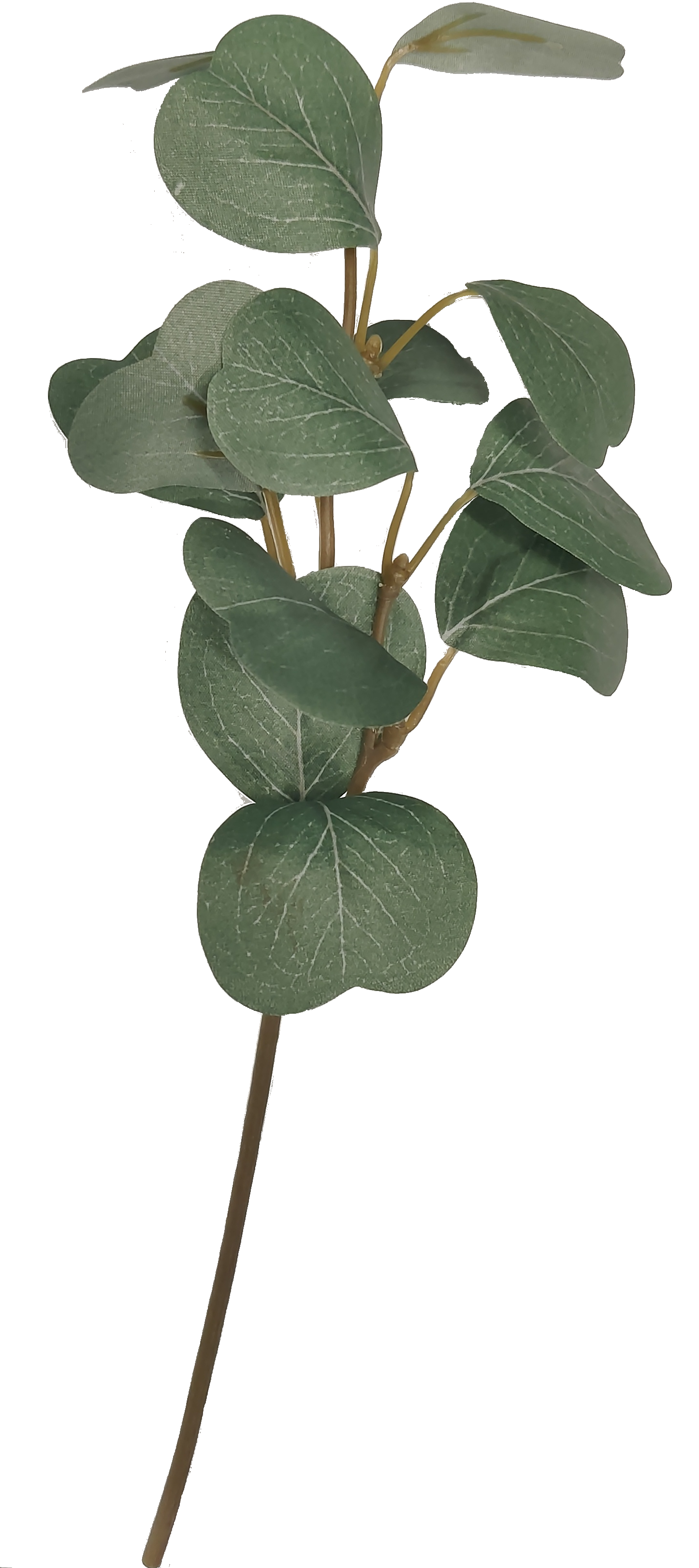}}
    \hfill
    \subfloat[]{
        \includegraphics[width=0.40\linewidth]{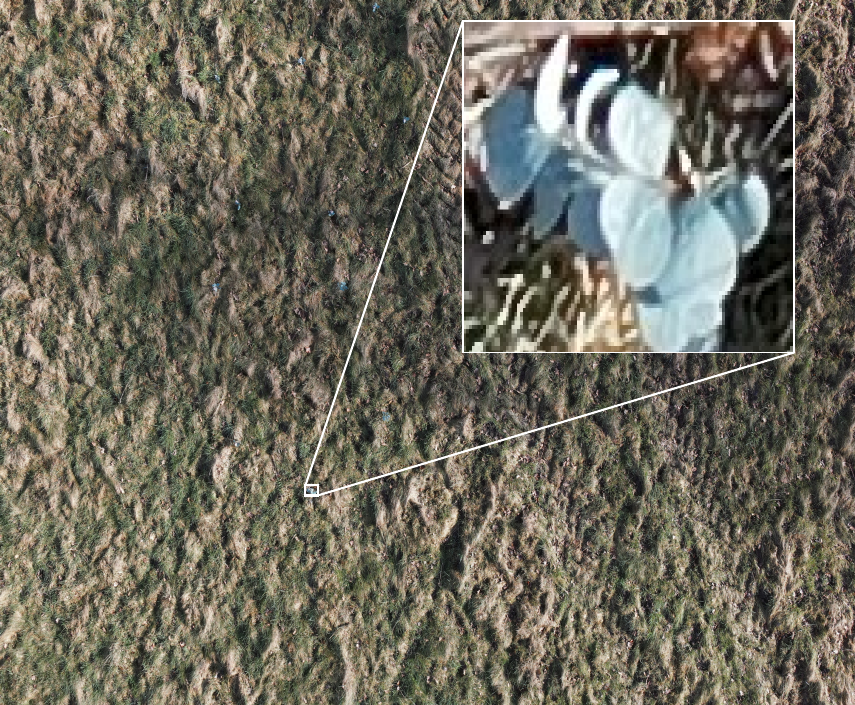}}
    \hfill
    \subfloat[]{
        \includegraphics[width=0.40\linewidth]{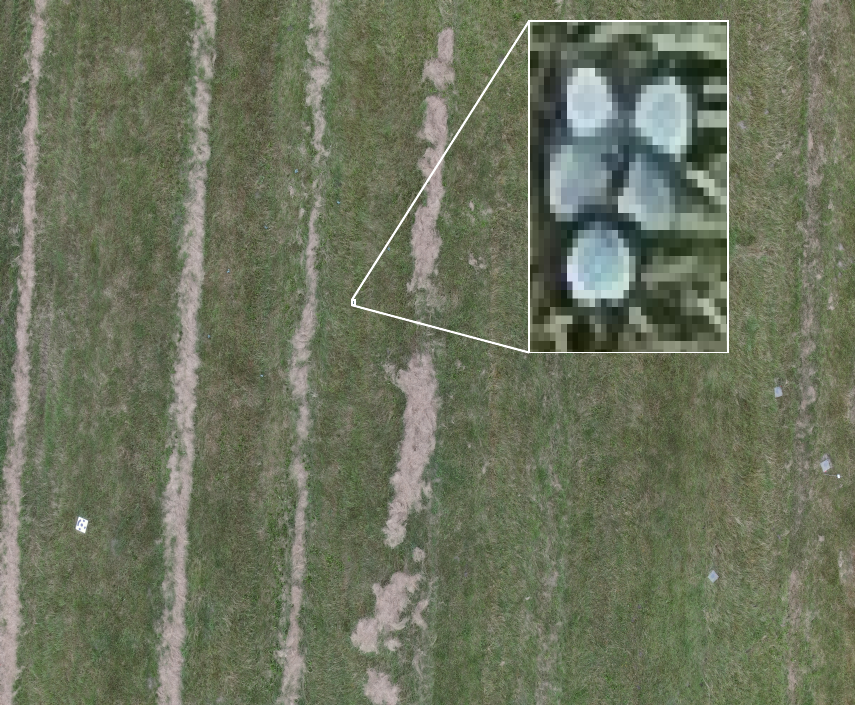}}
    \\
    \subfloat[]{
        \includegraphics[width=0.15\linewidth]{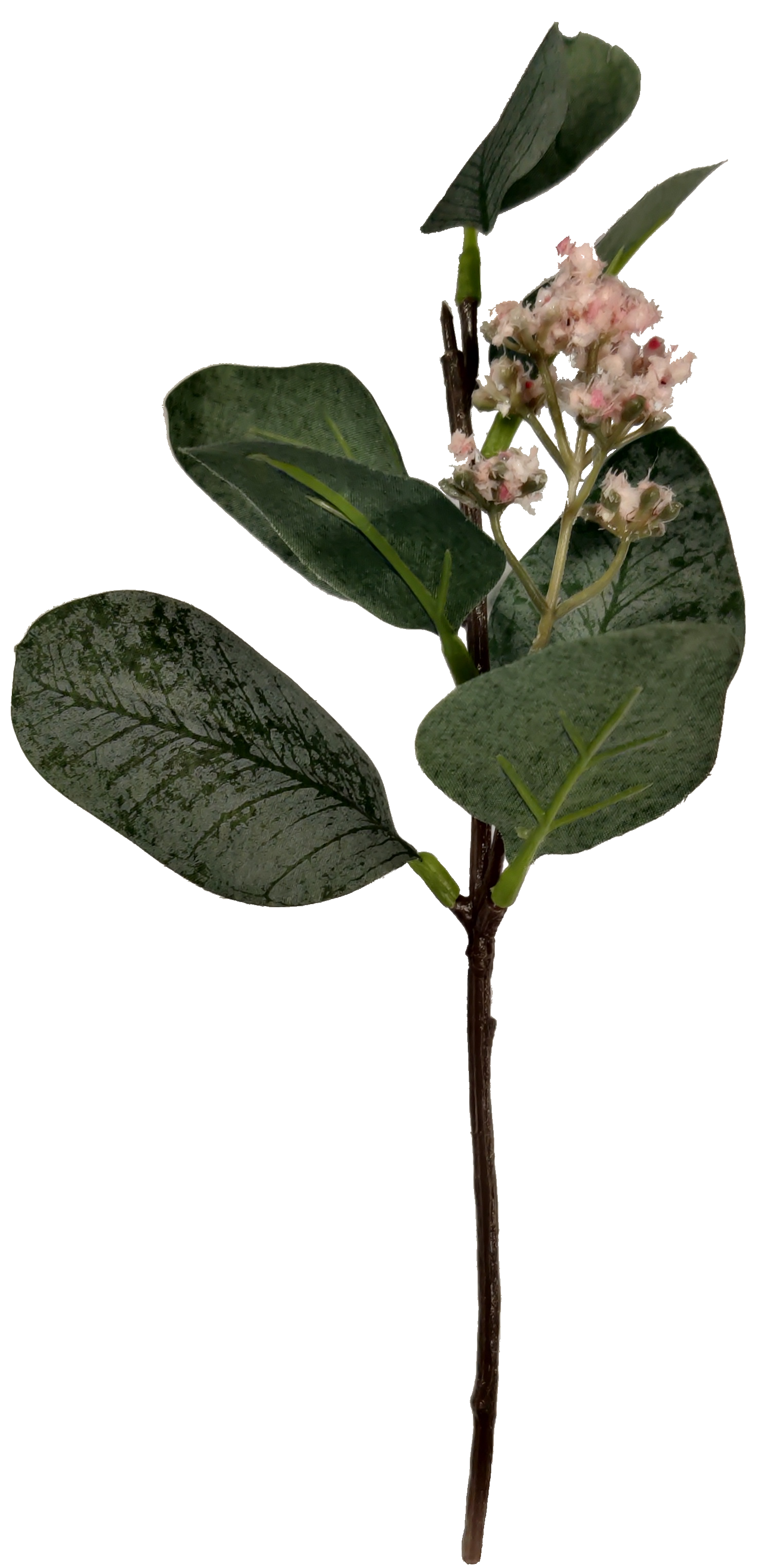}}
    \hfill
    \subfloat[]{
        \includegraphics[width=0.40\linewidth]{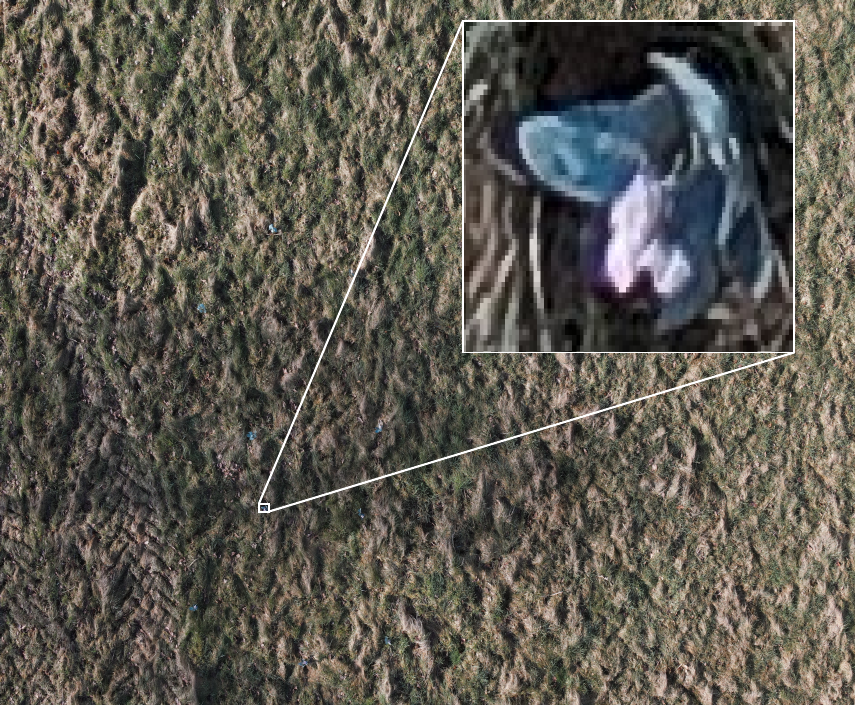}}
    \hfill
    \subfloat[]{
        \includegraphics[width=0.40\linewidth]{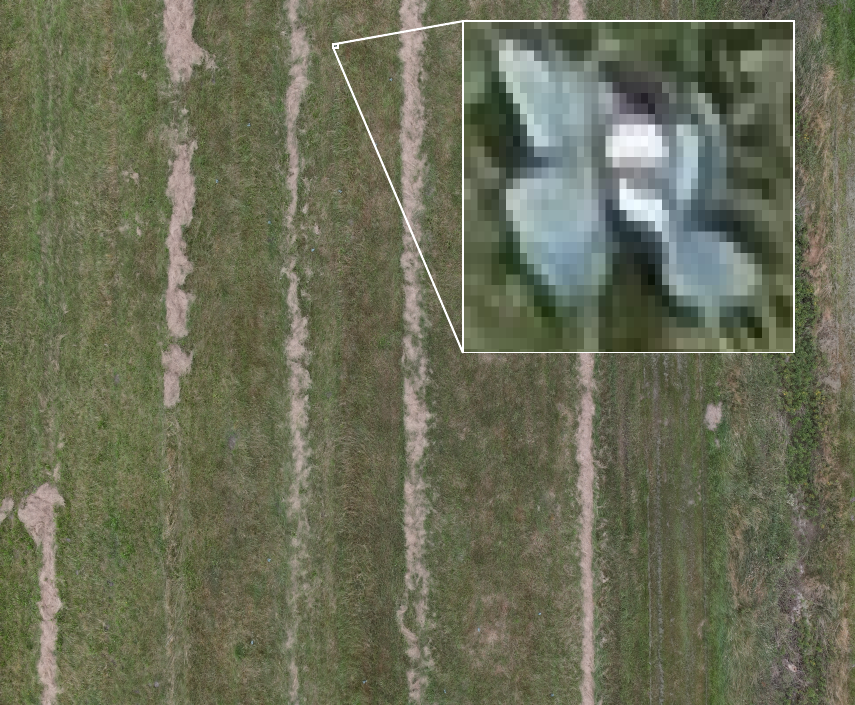}}
    \caption{Example of weed type I (a-c) and weed type II (d-f) at 12m (b, e) and 32m (c, f) altitude on the datasets of 13 February 2024 (b, c) and 1 August 2024 (e, f). Note the presence of a white flower for weed type II.}
    \label{fig:example_images_with_annotations} 
\end{figure}

Image data was collected from a grass field of around 0.75 hectares on the experimental field station of Wageningen University and Research near the city of Wageningen, The Netherlands. To allow repeatable measurements throughout the season, we simulate a weed detection application by distributing two different types of artificial plants on the grass field (Figure \ref{fig:example_images_with_annotations}). The size of these plants is approximately 10 by 10 cm (top-down), and the minimum distance between two plants was set to 1m. Images were taken on four different dates (13 February 2024, 23 April 2024, 18 July 2024, and 1 August 2024) to improve the generalization of the detection network. Due to the availability of some parts of the field on some dates, the field shape and area slightly differ between the datasets. 

\begin{figure}[!ht] 
    \centering
    \subfloat[]{
        \includegraphics[width=0.45\linewidth]{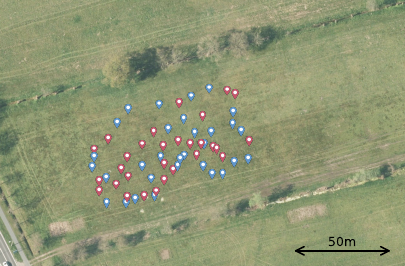}}
    \hfill
    \subfloat[]{
        \includegraphics[width=0.45\linewidth]{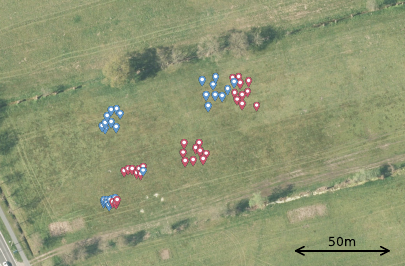}}
    \caption{The measured location using RTK of the artificial plants distributed in a uniform (a) and clustered (b) distribution on the dataset of 13 February 2024. The colors indicate the different weed types.}
    \label{fig:example_distributions} 
\end{figure}

Since weeds are non-uniformly distributed in fields \citep{Cardina1997} and the spatial distribution of weeds can have a large influence on the efficiency of an adaptive path planner \citep{vanEssen2024}, we investigate the effect of different weed distributions. At each date, two datasets were recorded by arranging the plants in two different distributions: a uniform distribution and a clustered distribution (Figure \ref{fig:example_distributions} shows the distributions over the field on the dataset of 13 February 2024). 

Images were acquired using a Matrice 300 RTK UAV equipped with a Zenmuse P1 camera (DJI, Shenzhen, China) with 35mm focal length. An overlap of 70\% side and 80\% forward was used with an image resolution of 8192x5460 pixels. Images were recorded at 12m, 24m, and 32m altitude. Additionally, eight ground control points (GCPs) were placed in the field to improve georeferencing during creation of the orthomosaics, further explained in section \ref{sec:offline_sim_env}. For evaluation, the location of the weeds and the GCPs were measured using a Topcon HiPer SR RTK-GNSS receiver (Topcon Positioning Systems, Inc., Tokyo, Japan) with around 1.0 - 1.5 cm accuracy.

\subsection{Training of the detection network}
\label{sec:detection_network_training}
For each date and distribution, 10\% of the plants (6 plants per date and distribution) were randomly assigned to the training set and 5\% to the validation set (3 plants per date and distribution). This leaves the majority of the plants unseen by the detection network, which was required for a fair evaluation of the adaptive path planner. Because each plant was visible in multiple images due to the overlap between the images, all images containing the training and validation plants were selected from the original UAV images at 12m, 24m, and 32m altitude and manually annotated. To prevent data leakage between the training and validation datasets, in the training dataset, all plants that were not part of the training set were masked out by replacing the pixel values within the bounding box with a green color. Likewise, in the validation dataset, we masked out all plants that did not belong to the validation set. The detection network was trained on all datasets together, combining all dates, altitudes, and distributions for generalization. The combined training dataset contained 1618 images with 2000 annotations, and the combined validation dataset contained 889 images with 977 annotations, respectively. 

An image input size of 2048x2048 was used for the detection network, all other hyperparameters were kept at their default value. The network was trained for 250 epochs with batch size of 16 using a computer with an AMD Ryzen 5950x CPU and NVIDIA RTX 3090 graphics card.

\subsection{Offline simulation environment}
\label{sec:offline_sim_env}
To be able to do a large number of experiments and parameter optimization with the adaptive planner, we used the captured images to create georeferenced orthomosaics in order to get a realistic offline simulation environment. For each date and plant distribution, an orthomosaic was made from the 12m altitude images using Agisoft Metashape v2.1.0 (Agisoft LLC, St. Petersburg, Russia). The eight ground control points were used to improve georeferencing. In total, eight different orthomosaics were made using the high-quality settings. 

For each waypoint, we calculated the field-of-view of the drone, rotated the orthomosaic using the UAV's heading and cropped the field-of-view from the orthomosaic around the GNSS location of the UAV. The resulting image was resized to the original image size using linear interpolation. The generated images looked visually similar to the original images taken by the UAV.

\subsection{Baseline}
\label{sec:baseline}
We compare the adaptive path planner with a coverage planner planned using Fields2Cover \citep{Mier2023}, covering the complete field at fixed altitude. Similarly, as for the adaptive path planner, the row width and distance between two waypoints were set to 90\% of the field-of-view. 

\subsection{Performance metrics}
\label{sec:performance_metrics}
Both path planners were evaluated using the detection performance and flight path length. The locations of mapped objects with a certainty larger than $c_\textrm{eval}=0.5$ were compared to the ground truth using their GNSS location measured in the field. A mapped object was counted as a true positive (TP) when the class was equal to the ground truth class and the Euclidean distance from the ground truth location was smaller than 0.35m. When a mapped object had no ground truth location, it was counted as a false positive (FP), when an object was not present in the map, it was counted as a false negative (FN). The detection performance was quantified using the F1-score. 

The path length was evaluated by summing the Euclidean distance between the waypoints. Since the area and shape of the datasets were not equal, we calculated the relative distance, $r_{k}^\textrm{diff}$, between the adaptive flight and the baseline coverage planner:

\begin{equation}
    r_{k}^\textrm{diff} = \frac{l_{k}}{l_k^\textrm{cov}},
\end{equation}
\noindent
where $l_{k}$ is the flight path length for the adaptive path planner on dataset $k$ and $l_k^\textrm{cov}$ is the length of the corresponding baseline coverage path at 12m altitude. When $r_{k}^\textrm{diff}$ is larger than 1, the adaptive flight path is longer than the baseline, and when $r_{k}^\textrm{diff}$ is smaller than 1, it is shorter than the baseline.

\subsection{Experiments}
\label{sec:experiments}
\subsubsection{Experiment 1: Relationship between detection certainty and errors}
The certainty measure of a detection is an important part of the adaptive path planner, as it determines whether a low-altitude inspection is needed. Therefore, in order to make meaningful decisions, the certainty measure needs to provide separation between true and false positives. To investigate the usability of the different certainty measures (defined in \ref{sec:unc_calc}) at different altitudes, we evaluated them on the captured images of 12m, 24m, and 32m altitude for the 8 datasets. Errors made within 0.35m of the image border were ignored because the ground truth object could be outside the image due to inaccuracies in both the image GNSS location and measured ground truth object location, which would result in an incorrect evaluation. For each certainty measure, we show the certainties for the TP and FP detections for the three altitudes. A large separability between the two indicates that the certainty measure can be used to differentiate between certain and uncertain detections. To quantify the separability between TP and FP detections using the certainty measure, we calculate the t-value for the alternative hypothesis that the mean confidence of the TPs, $\mu_\textrm{tp}$, is higher than the mean confidence of the false positive, $\mu_\textrm{fp}$, using Welch's t-test \citep{Welch1947}:

\begin{equation}
    t = \frac{\mu_\textrm{tp} - \mu_\textrm{fp}}{\sqrt{\Bigl( \frac{\sigma_\textrm{tp}}{\sqrt{N_\textrm{tp}}} \Bigr)^2 + \Bigl( \frac{\sigma_\textrm{fp}}{\sqrt{N_\textrm{fp}}} \Bigr)^2}},
\end{equation}
\noindent
where $N$ is the sample size and $\sigma$ the standard deviation. The higher the value of $t$, the better the certainty measure can be used to differentiate between true and false positives. The certainty measure with the largest t-value for each altitude was used in experiments 2, 3, and 4.

\subsubsection{Experiment 2: Parameter optimization}
The altitude of the coverage flight path ($h_{\textrm{cov}}$), the acceptance confidence threshold ($c_{\textrm{accept}}$), and the rejection confidence threshold ($c_{\textrm{reject}}$) are important parameters for the adaptive path planner. These parameters are dependent on each other; for example, a higher coverage altitude works best with a low rejection confidence, because the detection certainty is lower at a higher altitude, and a high rejection confidence will therefore reject most detections, resulting in many false negatives. To study the influence of these parameters on the performance of the adaptive path planner, we ranged $h_\textrm{cov}$ from 12m to 48m in steps of 12m, $c_\textrm{accept}$ from 1.0 to 0.4 in steps of 0.2 and $c_\textrm{reject}$ with the values of 0.05, 0.2 and 0.4. All combinations are examined, except the combination of $c_{\textrm{accept}}=0.4$ and $c_{\textrm{reject}}=0.4$ as this combination would never result in the decision $d_i$ to inspect a detection (see section \ref{sec:inspection_flight_path}). We evaluated the F1-score and the relative difference in path length with the coverage flight path for all possible combinations of these parameters using the path planner on the clustered and uniform orthomosaics. The set of parameters that had both a good detection performance and a short path length was used in experiments 3 and 4. For all experiments, the inspection altitude, $h_\textrm{inspect}$, was set to 12m to match the minimum altitude of the UAV used during data collection. For comparison, we also evaluated the F1-score and relative flight path length for the coverage path planner for the same coverage altitudes ($h_\textrm{cov}$).

\subsubsection{Experiment 3: Localization errors}
\label{sec:experiments_localization_error}
Errors in the localization of the drone position and orientation directly influence the georeferencing of the detected objects. Georeferencing errors can result from errors in the position and altitude estimates by the UAV's GNSS, and errors in estimates of roll, pitch, and heading by the UAV's IMU. To investigate the effect of the localization errors on the F1-score, we experimented with five levels of localization error. Table \ref{tab:localization_levels} defines the maximum position, altitude, gimbal roll, gimbal pitch, and UAV heading errors for the different levels. The errors are sampled from a uniform distribution. The DJI M300 UAV used for the data collection had a positional error of around 1.5cm horizontal and vertical in the experiment field (provided by manufacturer) and an error in roll, pitch and heading of around 0.5 degree (average georeferencing error during the orthomosaic creation), which matches level 'good' in Table \ref{tab:localization_levels}. The errors in position and orientation were implemented by altering the cropped location and heading in the offline simulation environment. Because a higher localization error causes mismatching during the mapping of objects resulting in the creation of new objects in the map instead of updating existing ones (section \ref{sec:mapping}), we increased the maximum distance between a mapped object and a detection, $d_\textrm{dist}$, to 0.5m, 0.9m, 1.8m, and 4.7m for the uncertainty levels good, decent, poor and very poor respectively. Level perfect keeps the default value of 0.35m. These values were computed by calculating the offset when all maximum errors, as defined in Table \ref{tab:localization_levels}, occur simultaneously. Note that these maximum distances were only increased for the mapping; during evaluation, we kept the maximum distance between a ground truth object and a detected object at the default value of 0.35m for a fair assessment. For each level, we compared the F1-score for the adaptive path planner with a coverage flight path at 12m, 24m, 36m and 48m altitude for both the clustered and uniform orthomosaics.

\begin{table}[t]
    \centering
    \caption{Localization uncertainty levels used in experiment 3 expressed as offset in drone position, altitude, and heading, and camera gimbal roll and pitch. The values are the minimum and maximum of the offset sampled from a uniform distribution.}
    \begin{tabular}{cccccc}
        \hline
         Level     & Position    & Altitude    & Roll                &  Pitch              & Heading             \\
        \hline
         Perfect   & $\pm0.000$m & $\pm0.000$m & $\pm0.0$\textdegree & $\pm0.0$\textdegree & $\pm0.0$\textdegree \\
         Good      & $\pm0.015$m & $\pm0.015$m & $\pm0.5$\textdegree & $\pm0.5$\textdegree & $\pm0.5$\textdegree \\
         Decent    & $\pm0.030$m & $\pm0.030$m & $\pm1.0$\textdegree & $\pm1.0$\textdegree & $\pm1.0$\textdegree \\
         Poor      & $\pm0.100$m & $\pm0.100$m & $\pm2.0$\textdegree & $\pm2.0$\textdegree & $\pm2.0$\textdegree \\
         Very poor & $\pm0.200$m & $\pm0.200$m & $\pm5.0$\textdegree & $\pm5.0$\textdegree & $\pm5.0$\textdegree \\
        \hline
    \end{tabular}
    \label{tab:localization_levels}
\end{table}

\subsubsection{Experiment 4: Number of objects}
To investigate the effect of different numbers of objects in the field on the flight path length, the number of objects in the field was altered. For each dataset, 140 additional objects were created by cropping annotated plants from the training dataset, to get a total of 200 objects. These cropped images were randomly rotated and added to the orthomosaic, creating realistic images for the detection network. Depending on the distribution in the original orthomosaic, the additional objects were distributed in the field using the following method:

\noindent
\textbf{Clustered distribution:} the number of objects in a cluster, $k$, was drawn from a normal distribution:

\begin{equation}
    k = \textrm{round}(x), x \sim \mathcal{N}(\mu=8, \sigma=3),
\end{equation}
\noindent
where $\mu$ is the mean and $\sigma$ the standard deviation. Inside each cluster, the objects were distributed using a multivariate Gaussian distribution with a random mean location in the field and a random positive semidefinite covariance matrix $\Sigma_\textrm{loc}$:

\begin{equation}
    \Sigma = \begin{bmatrix}
        a_{11} & a_{12} \\
        a_{21} & a_{22} \\
    \end{bmatrix}, a_{ij} \sim \mathcal{N}(\mu=5, \sigma=2),
\end{equation}
\begin{equation}
    \Sigma_\textrm{loc} = \Sigma \cdot \Sigma^T.
\end{equation}
\noindent
The minimal distance between cluster centers was set to 15.0m and the minimum distance between all objects was set to 1.0m, the location was resampled when these requirements were not met.

\noindent
\textbf{Uniform distribution:} the additional objects were uniformly distributed in the field. The minimum distance between two objects was set to 1.0m.

\begin{figure}[!ht] 
    \centering
    \subfloat[]{
        \includegraphics[width=0.45\linewidth]{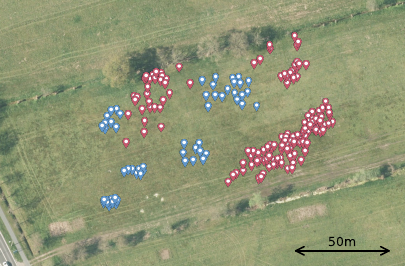}}
    \hfill
    \subfloat[]{
        \includegraphics[width=0.45\linewidth]{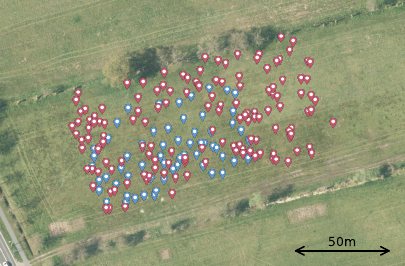}}
    \caption{Example of the original (blue) and added (red) objects on the dataset of 13 February 2024 for the clustered (a) and uniform (b) distribution.}
    \label{fig:example_additional_objects} 
\end{figure}

We tested the influence of the different number of objects by ranging the number of objects between 0 and 200 in steps of 20. Since each original orthomosaic contained 60 objects, we masked out object locations when we needed fewer than 60 objects by replacing the pixel values with a green color. When more than 60 objects were needed, we added them from the additional objects. For the clustered distribution, the object locations were sorted by cluster to make sure that we added or removed objects within a cluster first before adding or removing another cluster. Figure \ref{fig:example_additional_objects} shows an example of the original and added object locations on the dataset of 13 February 2024 for both the clustered and uniform distributions. For each number of objects, we compared the object density in the number of objects per hectare with the relative difference in flight path between the adaptive planner and the baseline. This was done for the orthomosaics with a clustered and uniform distribution of objects separately.

\section{Results}
\subsection{Experiment 1: Relationship between detection certainty and errors}
\begin{figure}[!ht]
   \centering
   \includegraphics[width=\textwidth]{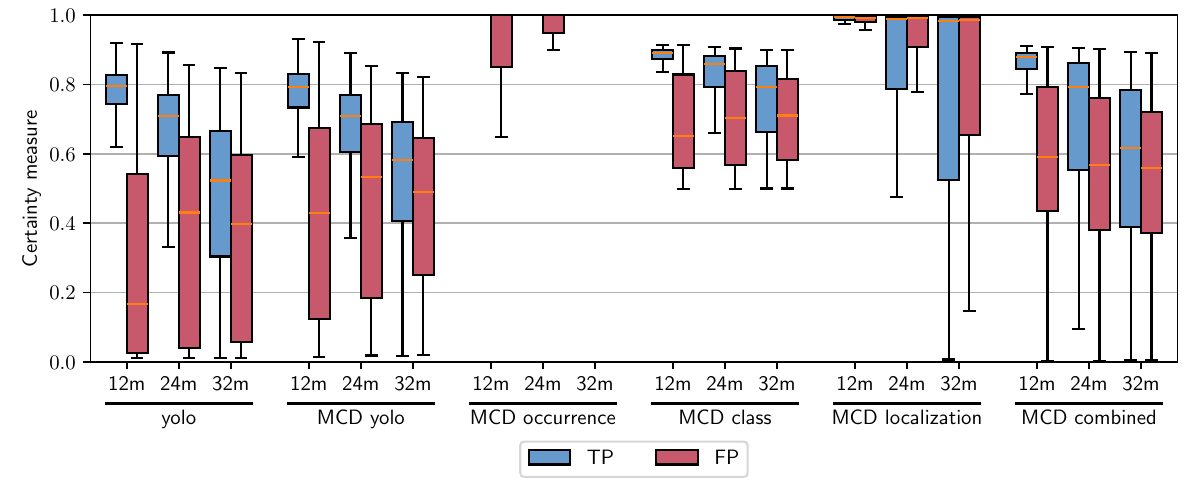}
   \caption{Boxplot showing the distribution of the true positive (TP) and false positive (FP) detections for the standard YOLOv8 certainty measures and the certainty measures derived using Monte-Carlo Dropout (MCD) at different altitudes.}
   \label{fig:result_uncertainty_boxplot}
\end{figure}

To investigate the ability of the different detection certainty measures to separate TP and FP detections, Figure \ref{fig:result_uncertainty_boxplot} shows the distribution for the YOLOv8 confidence value and the certainty measures obtained using MCD for the TP and FP detections at different altitudes. A good certainty measure should result in a clear distinction between TP detections with a high certainty and FP detections with a low certainty. The certainty for the TP detections was, on average, higher than for FP detections, except for the localization certainty at 24m and 32m altitude. For all certainty measures, an increasing altitude resulted in an increasing overlap between the certainty distributions of the TP and FP detections. The default YOLOv8 confidence value showed a clear distinction for TP and FP detections at 12m altitude; however, increasing the altitude increased the overlap between the TP and FP confidences. The MCD occurrence measure resulted in most cases in a certainty of 1.0, and therefore could not be used to separate TPs and FPs. The MCD class uncertainty measure showed a clear distinction between TPs and FPs at 12m altitude. The localization certainty showed a distinction between TP and FP detections for 12m altitude. However, the distributions at higher altitudes did show more overlap compared to the 12m altitude. The combined MCD certainty value did not show a better distinction between true and false positives than the default YOLOv8 confidence score.

Table \ref{tab:certainty_method_t_values} shows the t-values resulting from Welch's t-test for testing the alternative hypothesis that the mean confidence of true positives is higher than the mean confidence of the false positives. The higher the t-value, the better the certainty measure is in separating true and false positives. As could also be seen in Figure \ref{fig:result_uncertainty_boxplot}, an increasing altitude decreased separability between the TP and FP distributions, indicating a more difficult distinction between correct and wrong detections. The standard YOLOv8 certainty measure performed best at 12m and 24m altitude. The MCD class performed best at 32m altitude, with the default YOLOV8 confidence score as second-best. 

On average, applying MCP during inference resulted in a total processing time for a single image including MCD (preprocessing, inference and postprocessing) of $105.2\pm116.1$ ms compared to a total processing time of $22.1\pm2.9$ ms for YOLOv8 without dropout. 

Using the MCD certainty measures did not result in a better separation between correct and wrong detections for 12m and 24m, and the difference in performance between the MCD class and the YOLOv8 confidence measure at 32m was small. Additionally, applying MCD increased the processing time by a factor of 5. Therefore, the rest of the experiments use the YOLOv8 confidence value as the certainty measure. The average F1-scores for all images in the dataset, using the YOLOv8 confidence value as the certainty measure, were 0.83 at 12m, 0.70 at 24m, and 0.44 at 32m altitude. 

\begin{table}[t]
    \centering
    \caption{T-values from Welch’s t-test for standard yolo-v8 certainty and the certainty methods using Monte-Carlo Dropout (MCD), testing the alternative hypothesis that the mean confidence of true positives is higher than the mean confidence of the false positives. Values indicated with a '*' show a significant ($\alpha=0.05$) difference between the true and false positive distributions; the highest t-values for each altitude are indicated in bold.}
    \begin{tabular}{cccc}
        \hline
         Certainty method & 12m             & 24m            & 32m            \\
        \hline
         yolo             & \textbf{101.1*} & \textbf{48.3*} & 18.0*          \\
         MCD yolo         & 77.8*           & 40.7*          & 15.6*          \\
         MCD occurrence   & 41.6*           & 25.6*          & 11.1*          \\
         MCD class        & 78.7*           & 42.5*          & \textbf{18.3*} \\
         MCD localization & 3.4*            & -3.2           & -5.1           \\
         MCD combined     & 61.4*           & 23.1*          & 5.7*           \\
        \hline
    \end{tabular}
    \label{tab:certainty_method_t_values}
\end{table}

\subsection{Experiment 2: Parameter optimization}
\begin{figure}[!ht]  
    \centering
    \subfloat[]{
        \includegraphics[width=0.49\linewidth]{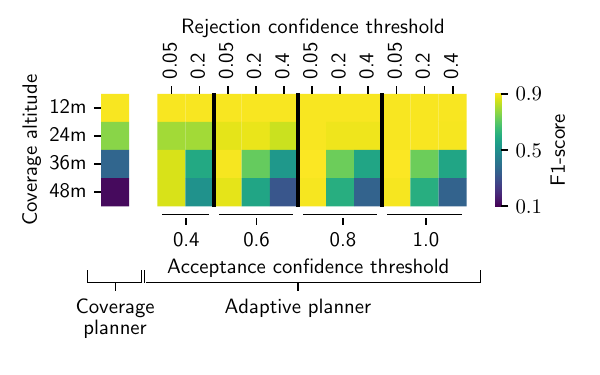}}
    \hfill
    \subfloat[]{
        \includegraphics[width=0.49\linewidth]{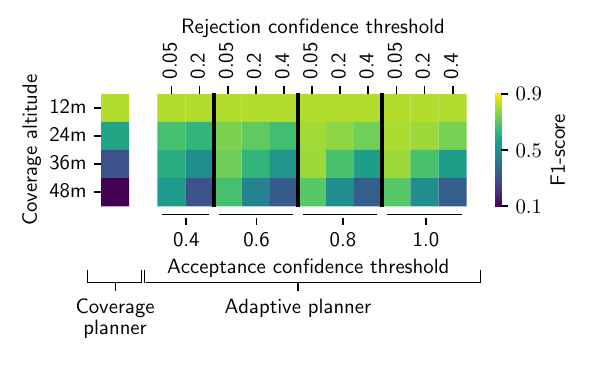}}
    \caption{Heatmap showing the F1-score of different values for the coverage altitude, the acceptance confidence threshold, and rejection confidence threshold on the F1-score for the coverage and the adaptive planner on the clustered (a) and uniform (b) distributed datasets.}
    \label{fig:result_parameter_estimation_f1} 
\end{figure}

Figure \ref{fig:result_parameter_estimation_f1} shows the influence of the coverage altitude ($h_\textrm{cov}$), acceptance confidence threshold ($c_\textrm{accept}$) and rejection confidence threshold ($c_\textrm{reject}$) on the F1-score for the adaptive planner and the coverage planner for both the clustered and the uniformly distributed plants. For the coverage planner, the F1-score decreased when the altitude increased. For the adaptive planner, the performance also decreased with an increasing altitude, however, to a lesser extent than the coverage planner, depending on the rejection confidence threshold. Increasing the rejection confidence lowered the F1-score for both distributions, especially in combination with a high coverage altitude. Overall, the F1-scores were higher on the clustered distribution compared to the uniform distribution, however, the effect of changing the parameters was similar for both distributions.

\begin{figure}[!ht]  
    \centering
    \subfloat[]{
        \includegraphics[width=0.49\linewidth]{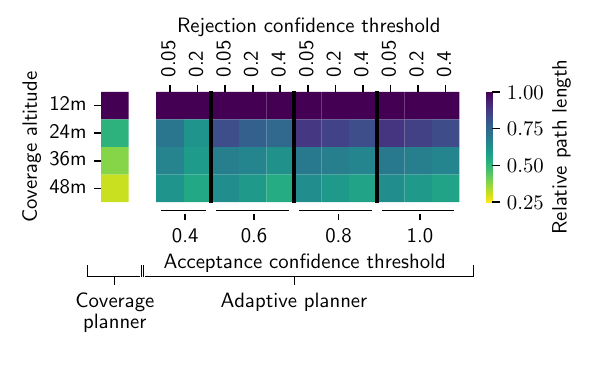}}
    \hfill
    \subfloat[]{
        \includegraphics[width=0.49\linewidth]{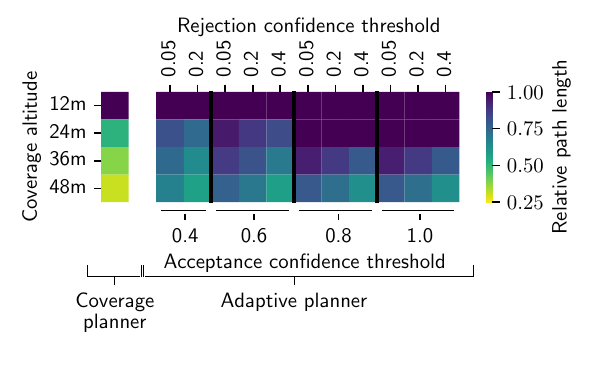}}
     \caption{Heatmap showing the relative flight path length of different values for the coverage altitude, the acceptance confidence threshold, and rejection confidence threshold on the relative path length for the coverage and adaptive planner on the clustered (a) and uniform (b) distributed datasets.}
    \label{fig:result_parameter_estimation_distance} 
\end{figure}

Figure \ref{fig:result_parameter_estimation_distance} shows the influence of the coverage altitude, acceptance confidence threshold, and the rejection confidence threshold on the relative path length (with respect to a coverage planner at 12m altitude) for the coverage and adaptive planner. Flying at 48m altitude decreases the path length for the coverage planner by around 70\%. The adaptive planner yielded a longer flight path than the coverage planner at 48m altitude due to the addition of the inspection flight path; however, it is still around 40\% shorter than the 12m coverage flight path. Increasing the rejection confidence decreased the flight path length due to a lower number of objects that needed to be inspected at a lower altitude. This effect was stronger for the uniform distribution, since the distance between the objects was on average larger than for the clustered distribution. A higher acceptance threshold resulted in a longer flight path, because more objects needed to be inspected at low altitude. However, this effect was only visible at a coverage altitude of 24m.

Figures \ref{fig:result_parameter_estimation_f1} and \ref{fig:result_parameter_estimation_distance} together show the trade-off between detection performance and flight path length. Using the coverage altitude of 12m resulted in the highest F1-score, while at the same time also gave the longest flight path length. Having a high rejection confidence at 48m coverage altitude resulted in a high number of FNs and thereby limited the number of low-altitude inspections, which resulted in a short flight path length and a low detection performance. Alternatively, a low rejection confidence threshold resulted in a higher detection performance, but at the cost of a longer flight path length.

\begin{figure}[!ht]  
    \centering
    \includegraphics[width=\textwidth]{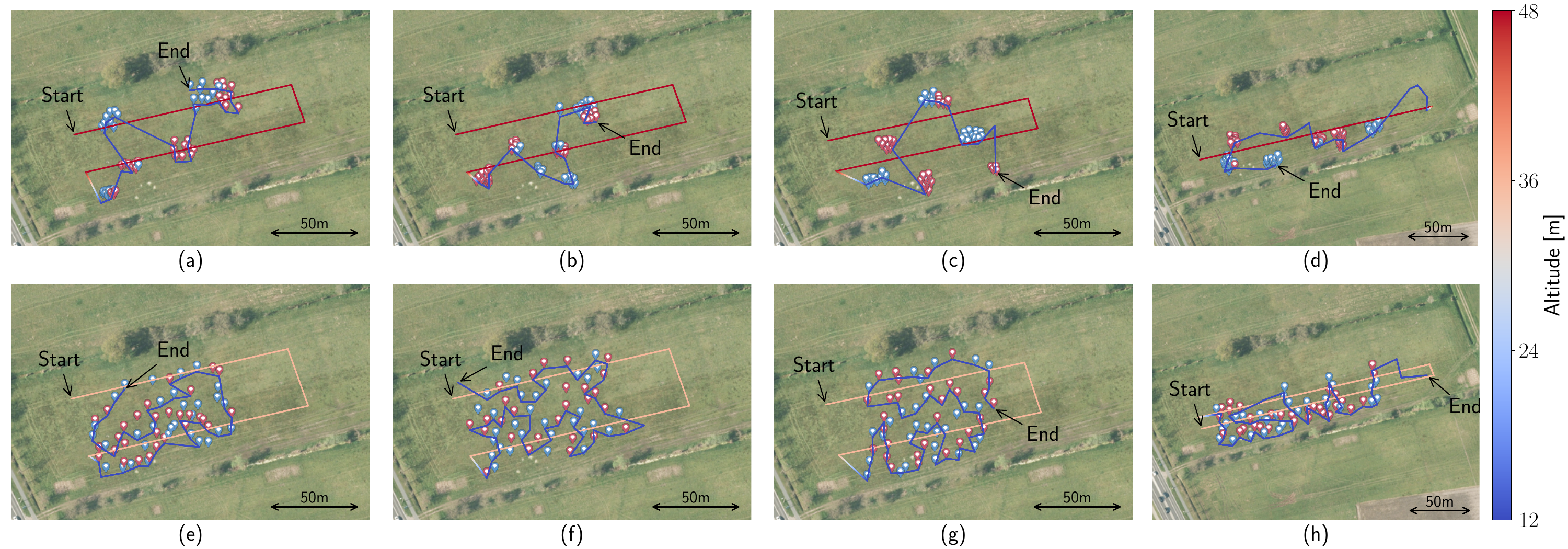}
    \caption{Flight paths for the adaptive planner using the optimized parameters corresponding to the datasets of 13 February 2024 (a, e), 23 April 2024 (b, f), 18 July 2024 (c, g) and 1 Augustus 2024 (d, h) with a clustered (a-d) and uniform (e-h) distribution of plants. The start and end location of the UAV are indicated with an arrow; the ground truth locations of weed types I and II are indicated with red and blue markers, respectively.}
   \label{fig:result_parameter_estimation_flight_path}
\end{figure}

Table \ref{tab:best_parameters} shows the parameters that yielded both a good performance and a relatively short flight path length. These parameters were used in experiments 3 and 4. Additionally, it shows the corresponding F1-score for the adaptive planner, the coverage planner using the same coverage altitude, and the normalized flight path length. The coverage flight path has a higher F1-score than the adaptive flight path for both the clustered distribution and the uniform distribution, however, at the cost of a longer flight path length. At the cost of a 2\% lower F1-score, the flight path length decreased by 37\% for the clustered distribution. For the uniform distribution, the advantages of the adaptive path planner are smaller, at the cost of a 2\% lower F1-score, the flight path length only decreased 6\%. Figure \ref{fig:result_parameter_estimation_flight_path} shows the resulting adaptive flight path for both the clustered and uniform distribution of objects on the four different recording dates.

\begin{table}[t]
    \centering
    \caption{Selected parameters and the resulting F1-score and normalized flight path length for the clustered distribution and the uniform distribution, having a high F1-score and a short flight path length.}
    \begin{tabular}{ccc}
        \hline
         Parameter                     & Clustered distribution & Uniform distribution \\
        \hline
         Coverage altitude             & 48m                    & 36m                  \\
         Acceptance threshold          & 0.6                    & 0.8                  \\
         Rejection threshold           & 0.05                   & 0.05                 \\
        \hline
         F1-score adaptive planner     & $0.867\pm0.024$        & $0.784\pm0.075$      \\
         F1-score coverage planner     & $0.891\pm0.021$        & $0.807\pm0.039$      \\
         Normalized flight path length & $0.631\pm0.035$        & $0.938\pm0.116$      \\
        \hline
    \end{tabular}
    \label{tab:best_parameters}
\end{table}
 
\subsection{Experiment 3: Localization errors}
\begin{figure}[!ht] 
    \centering
    \subfloat[]{
        \includegraphics[width=0.9\linewidth]{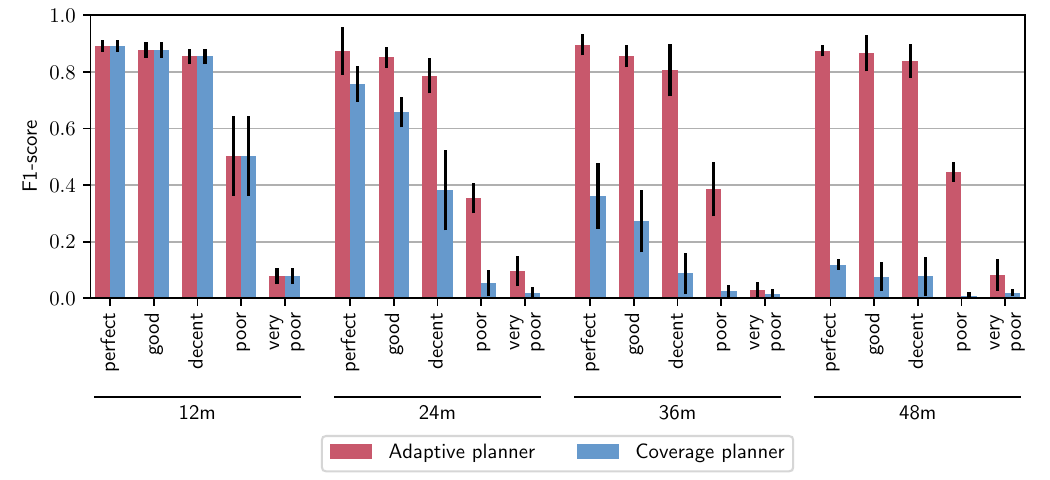}}
    \\    
    \subfloat[]{
        \includegraphics[width=0.9\linewidth]{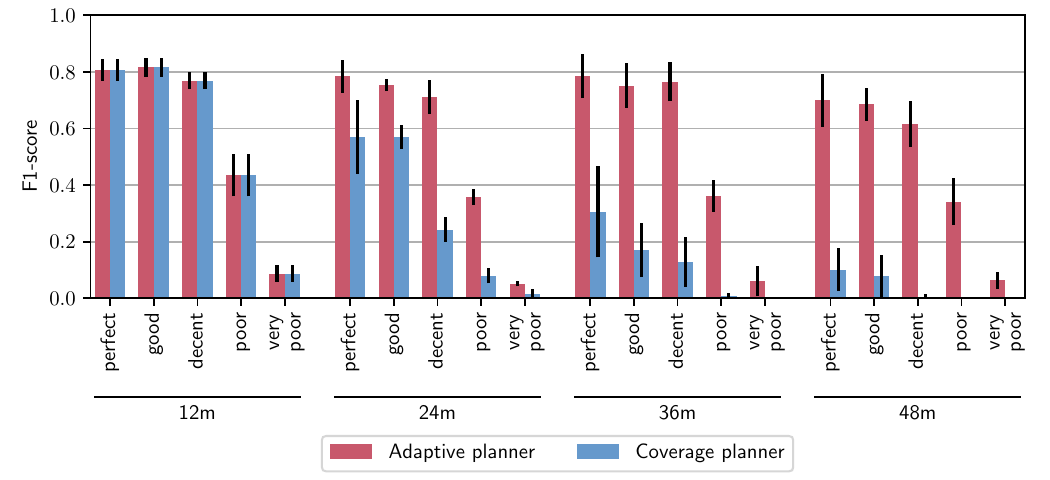}}
     \caption{Effect of localization error on the F1-score for the adaptive planner and coverage path planner at different altitudes shown for the clustered (a) and uniform (b) distribution of plants. The error bar indicates the standard deviation from the mean.}
    \label{fig:result_localization_uncertainty} 
\end{figure}

Figure \ref{fig:result_localization_uncertainty} shows the effect of localization errors on the performance of the adaptive planner and the low-altitude coverage planner for both the clustered and uniform distribution of objects. Five levels of error were simulated, from 'perfect' to 'very poor' (see section \ref{sec:experiments_localization_error}). A higher localization error resulted in a lower F1-score for both planners. However, since the absolute error in localization is smaller at a lower altitude, the adaptive path planner yielded a higher F1-score at 24m, 36m, and 48m altitude than the coverage flight path, as it used additional low-altitude inspections, thereby improving the object localization. There was a notable drop in performance between the levels 'decent' and 'poor', indicating a maximum allowable error of 0.03m in position and altitude, along with a 1-degree error in the UAV's heading, gimbal roll, and pitch. With a higher localization error, the low F1-scores can be explained by a higher number of FPs due to multiple observations of the same plant not being fused together, resulting in multiple plants. There was no notable difference in the F1-score between the clustered and uniform distribution of plants.

\subsection{Experiment 4: Number of objects}
\begin{figure}[!ht]
    \centering
    \subfloat[]{
        \includegraphics[width=0.49\linewidth]{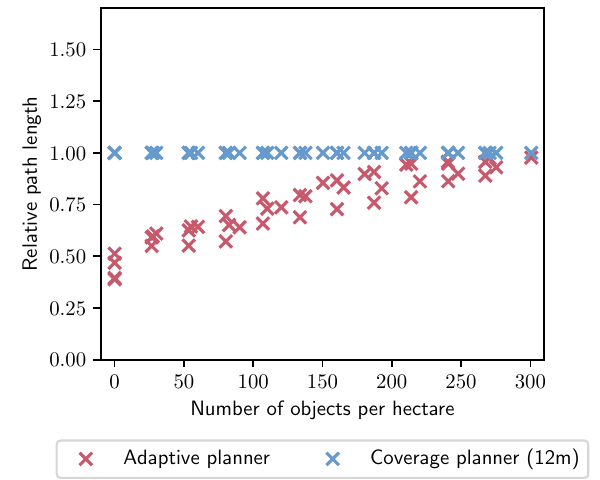}}
    \subfloat[]{
        \includegraphics[width=0.49\linewidth]{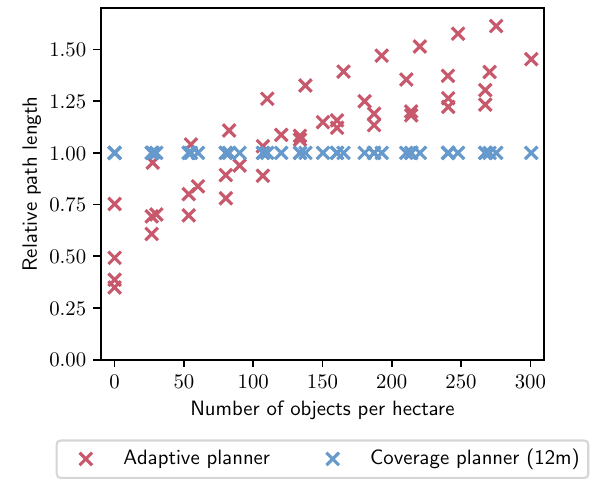}}
     \caption{Effect of the number of objects on the normalized flight path length for the adaptive planner and the coverage planner for the clustered (a) and uniform (b) distribution of plants.}
    \label{fig:result_number_of_objects} 
\end{figure}

Figure \ref{fig:result_number_of_objects} shows the effect of the number of objects in the field on the relative flight path length for the clustered and uniformly distributed objects. For a clustered distribution of objects, the adaptive planner resulted in a shorter flight path than the low-altitude coverage planner for all tested numbers of objects. However, when there were more than 250 objects per hectare, the adaptive flight path length was almost similar. For the uniform distribution of plants, however, the adaptive planner yielded a longer flight path when there were more than 80 - 120 objects per hectare in the field compared to the low-altitude coverage planner. With a more uniform object distribution, the adaptive planner has to inspect so many plants at low altitude that it is more efficient to use a full-coverage low-altitude flight path.

\begin{figure} 
    \centering
    \includegraphics[width=\textwidth]{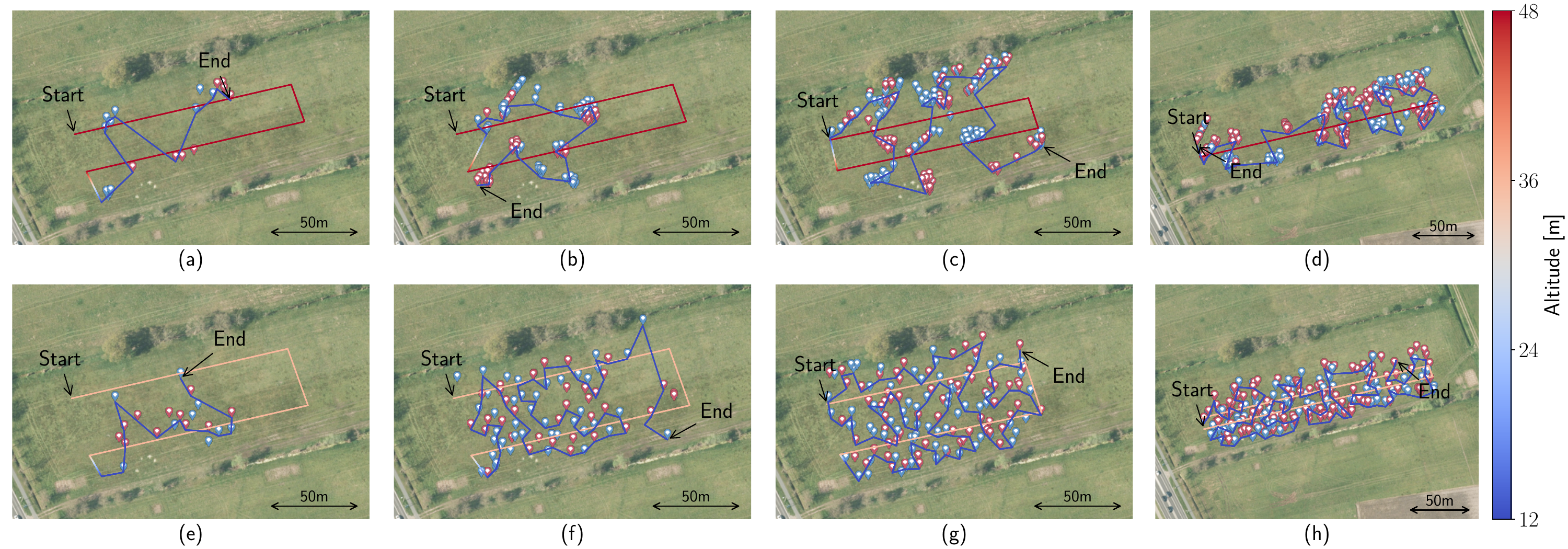}
    \caption{Flight paths for the adaptive planner corresponding to the datasets of 13 February 2024 with 20 objects (a, e), 23 April 2024 with 80 objects (b, f), 18 July 2024 with 140 objects (c, g) and 1 Augustus 2024 with 200 objects (d, h) with a clustered (a-d) and uniform (e-h) distribution of plants. The start and end location of the UAV are indicated with an arrow; the ground truth locations of weed types I and II are indicated with red and blue markers, respectively.}   \label{fig:results_number_of_objects_flight_path}
\end{figure}

Figure \ref{fig:results_number_of_objects_flight_path} shows the flight paths for the adaptive path planner and the coverage planner for the adapted dataset for the four different recording dates with 20, 80, 140, and 200 objects in the field. When the objects were uniformly distributed, the UAV had to visit locations that were more uniformly distributed over the field, requiring more waypoints and thereby increasing the flight path length compared to the clustered distribution.

\section{Discussion}
The results showed that, especially when the objects are distributed in clusters in the field, there is a lot of potential in using an adaptive path planner compared to a coverage path planner. Using an adaptive path planner with the optimized parameters, as defined in Table \ref{tab:best_parameters}, decreased the flight path length by 37\% on our datasets while the accuracy only decreased 2\% compared to the baseline when the objects were distributed in clusters. Compared to a coverage planner, the adaptive planner was more robust against localization errors (Figure \ref{fig:result_localization_uncertainty}). When increasing the number of objects in the field and the objects were distributed in clusters, the adaptive path planner outperformed a coverage path planner at 12m altitude till around 300 objects per hectare (Figure \ref{fig:result_number_of_objects}a). When the objects were uniformly distributed, however, the coverage planner outperformed the adaptive planner when there were more than 100 objects in the field  (Figure \ref{fig:result_number_of_objects}b). In section \ref{sec:comp_literature} we discuss the results in context of existing literature, section \ref{sec:disc_certainty_estimation} the detection certainty estimation, in section \ref{sec:disc_impact_dt_errors} the impact of the detection errors on the adaptive planner, section \ref{sec:disc_impact_loc_errors} the impact of the localization errors, section \ref{sec:disc_offline_sim_method} the validity of the offline simulation method and finally, section \ref{sec:disc_usability} discusses the usability of the adaptive planner in practice. 

\subsection{Comparison with literature}
\label{sec:comp_literature}
The results in this paper were compared to a full coverage flight path, which is the most used approach in practice, where it yielded a 37\% shorter flight path at the cost of a 2\% lower F1-score. \citet{vanEssen2024} showed a 74\% shorter flight path for a learning-based adaptive path planner, at the cost of a 12\% lower detection accuracy. \citet{Popovic2017} showed that their path planner detects 85\% of the objects in half the time of a coverage planner, while detecting all objects in approximately the same time as the coverage planner. They both show the trade-off between the flight time and the detection accuracy. Compared to those planners, our planner had a relatively small drop in performance, but also a smaller decrease in flight path length compared to a coverage flight. However, by selecting a different coverage altitude and confidence thresholds, it is possible to shift the balance towards a shorter flight path (experiment 2). 

Using the adaptive path planner only had advantages when the objects were non-uniformly distributed. This was also confirmed by \citet{Popovic2017} and \citet{vanEssen2024}, who both showed that the flight path was less efficient when the objects were uniformly distributed.

\subsection{Detection certainty estimation}
\label{sec:disc_certainty_estimation}
Using Monte-Carlo Dropout for certainty estimation did not show a better separability between TPs and FPs in experiment 1, except at 32 meter altitude. In the context of active learning, applying certainty estimation using MCD has been shown to estimate the most uncertain image for annotation \citep{Blok2022, Sokolova2024}, and has shown a positive correlation between certainty and detection accuracy. This could also be observed in experiment 1, where the TPs, on average, had a higher certainty than the FPs. In contrast to experiment 1, \citet{Myojin2019} observed a small increase in the F1-score when applying MCD uncertainty estimation to an adapted YOLOv3 network for lunar crater detection. However, the improvement was minimal, and since YOLOv8 is generally more accurate than YOLOv3 \citep{Jegham2024}, its confidence estimations are also more reliable.

A major disadvantage of sampling uncertainty using Monte-Carlo dropout is the added inference time. In our experiments, it increased the inference time by a factor of five, even with reusing intermediate results by the network. For the adaptive path planner, this could have an impact on the efficiency, especially when using onboard processing with limited computational power. Using alternative methods for uncertainty calculation that are less computationally expensive, such as Laplacian approximation \citep{Gawlikowski2023}, might be considered for future work.

\subsection{Impact of detection errors}
\label{sec:disc_impact_dt_errors}
Flying at a higher altitude increased the number of FN's for the detection network and thereby resulted in a lower F1-score (Figure \ref{fig:result_parameter_estimation_f1}). When there is an FN, no low-altitude inspection could be planned, and therefore, these objects were missed. Therefore, in experiment 2, the best working rejection confidence was the lowest of the tested values, 0.05, to have a minimum number of FNs. This resulted in more FP detections, however, they were corrected later on due to the low-altitude inspection. The impact of FN detections is stronger for the uniform distribution than for the clustered distribution, because for the clustered distribution, there will probably be a low-altitude inspection of some objects in the cluster. As long as some of the objects in the cluster are detected with a low detection certainty, most objects in the cluster will be detected by the low-altitude inspections, and also the ones that were missed from the high altitude. Experiment 2 indicated this by a lower optimal coverage altitude for the uniform distribution, which limited the number of FN's during the high-altitude coverage path. 

\subsection{Impact of localization errors}
\label{sec:disc_impact_loc_errors}
Experiment 3 showed that the adaptive planner better deals with localization errors than the baseline coverage planner,  till a position error of around 3 cm and an orientation error of around 1 degree. Because the localization error depends on the altitude, the low-altitude inspections decreased the localization error. Using a UAV equipped with RTK-GNSS should be accurate enough, indicating the practical usability of the planner. Additionally, the required accuracy of the localization also depends on the application. For instance, in a weed detection application, the resulting map may be used to send a weeding robot to a specific spot in the field. In that case, the required localization accuracy may be lower, since it is only needed to guide the weeding robot until the weeds are within its field-of-view.

\subsection{Validity of the offline simulation method}
\label{sec:disc_offline_sim_method}
The adaptive path planner was evaluated using a simulated flight over the orthomosaics created based on real drone images. Since orthomosaics are corrected for perspective distortion, the resulting images may simplify real-world images slightly. However, since the plants were much smaller than the flight altitude of the drone, taking snapshots of parts of the orthomosaic resulted in a similar perspective as during a real-world flight. The F1-scores on the real images in experiment 1 at 12m altitude were comparable with the F1-scores from the coverage planner at 12m altitude from the simulated images in experiments 2, 3, and 4. Additionally, the typical object detection performance in agriculture ranges from 0.7 to 0.9 in F1-score \citep{Rai2023, Ruigrok2023, Rehman2024}, which is comparable with the detection performance observed in our experiments. This indicates that the orthomosaic provides an accurate simulation of images taken in a real-world flight. Future work will focus on applying the path planner in the real world. 

\subsection{Operational opportunities for an adaptive path planner}
\label{sec:disc_usability}
The usability of the adaptive path planner depends on the application. When the objects are uniformly distributed in the field, a traditional low-altitude coverage flight path is efficient, except when there are less then 70 objects per hectare. However, when the objects are non-uniformly distributed, the adaptive path planner reduces the flight time with an comparable accuracy. For tasks such as weed mapping, using the adaptive path planner could be beneficial, as some weed species grow in distinct clusters in the field \citep{Cardina1997,Xu2023a}.

By increasing the flight speed of the UAV at higher altitudes, the advantages of an adaptive planner could be larger. At a higher altitude, the field-of-view of the UAV is larger. Since the flight speed depends on the motion blur, it is possible to increase the flight speed at larger altitudes \citep{Seifert2019}. This will result in a higher efficiency of the adaptive planner compared to a low-altitude coverage planner, as shown in this paper.

In this work, we used the same camera at high and low altitude; however, implementing multi-modal sensing could further increase the efficiency of the method. For example, detection of biotic stress of plants caused by infections can be done using thermal images \citep{Pineda2020}, although these cameras have a low spatial resolution, limiting their field-of-view. By combining a thermal camera with the RGB camera using the adaptive path planner, suspicious areas can be identified at high altitude using a high-resolution RGB camera, allowing for targeted closer inspection with a thermal camera at those specific spots.

\section{Conclusion}
In this study, we presented an adaptive path planning method that combines a high-altitude coverage path with low-altitude inspections based on the detection certainty. The path planner was evaluated on real-world orthomosaics derived from drone-based RGB images. Different detection certainty measures were compared, showing the standard YOLOv8 confidence value to be the best in differentiating true and false positive detections, thus most fit to be used in the adaptive path planner to determine the low-altitude inspection of detected, but uncertain plants. When the plants were distributed in clusters, the adaptive planner yielded a 37\% shorter flight path compared to a low-altitude coverage planner at the cost of only 2\% lower F1-score. When the objects were uniformly distributed in the field, the adaptive path planner showed a 6\% shorter flight path at the cost of a 2\% lower F1-score. The adaptive path planner could better deal with errors in the UAV's localization than a coverage planner by performing low-altitude inspections. When the objects were uniformly distributed, a coverage path was more efficient, except when there were fewer than 80-120 objects per hectare in the field. However, when the objects were non-uniformly distributed, the adaptive path planner was faster than a low-altitude coverage path till 250-300 objects per hectare. In conclusion, the presented adaptive path planner allowed finding non-uniformly distributed objects in a field faster than a coverage path planner and resulted in a comparable detection accuracy. 

\section*{CRediT author statement}
\textbf{Rick van Essen:} Conceptualization, Methodology, Formal analysis, Software, Visualization, Writing - Original Draft. \textbf{Eldert van Henten:} Conceptualization, Funding acquisition. \textbf{Lammert Kooistra:} Resources, Writing - Review \& Editing. \textbf{Gert Kootstra:} Conceptualization, Methodology, Writing - Review \& Editing, Funding acquisition.
 
\section*{Declaration of Competing Interest}
This research is part of the research program Synergia, funding was obtained from the Dutch Research Council (NWO grant 17626), IMEC-One Planet and other private parties. The authors have declared that no competing interest exist in the writing of this publication.

\section*{Data availability}
The orthomosaics, training images and network weights are made available at \url{https://doi.org/10.4121/bbe97051-07df-4934-b634-701d91a2075e} and the code at \repositoryURL.

\bibliographystyle{elsarticle-harv} 
\bibliography{references}

\end{document}